\pdfoutput=1

\documentclass[11pt]{article}
\usepackage[T1]{fontenc}
\usepackage[utf8]{inputenc}
\usepackage{listings}
\usepackage{xcolor}
\usepackage{longtable}
\usepackage{inconsolata}
\usepackage{microtype}

\usepackage{emnlp2023}

\usepackage{times}
\usepackage{latexsym}

\usepackage{float}
\restylefloat{table}

\usepackage[T1]{fontenc} 

\usepackage[utf8]{inputenc}

\usepackage{color-edits}
\addauthor{gn}{magenta}
\addauthor{byl}{purple}

\usepackage{microtype}
\usepackage{graphicx}
\graphicspath{{image/}}
\usepackage{pdfpages}
\usepackage{url}
\usepackage{hyperref}
\usepackage{xcolor}
\usepackage{epsfig}
\usepackage{adjustbox}
\usepackage{amsfonts}
\usepackage{amsmath}
\usepackage{amssymb}
\usepackage{booktabs} 
\usepackage{comment}
\usepackage{multirow}
\usepackage{caption}
\usepackage{subcaption}
\usepackage{textcomp}
\usepackage{comment}
\usepackage{relsize}
\usepackage{stmaryrd}
\usepackage{bbm}
\usepackage{rotating}
\usepackage{arydshln}
\usepackage{pifont}
\usepackage{xcolor}
\usepackage{colortbl}
\usepackage{tabularray}
\usepackage{graphicx}
\usepackage{pbox}
\usepackage[most]{tcolorbox}
\usepackage{colortbl}

\newtcolorbox{mybox}[2][]{
    colback=white,
    colframe=green!45,
    fonttitle=\bfseries,
    coltitle=black,
    sharp corners,
    title=#2,
    #1
}
\title{\raisebox{-0.1cm}{\includegraphics[width=0.500cm, height=0.637cm]{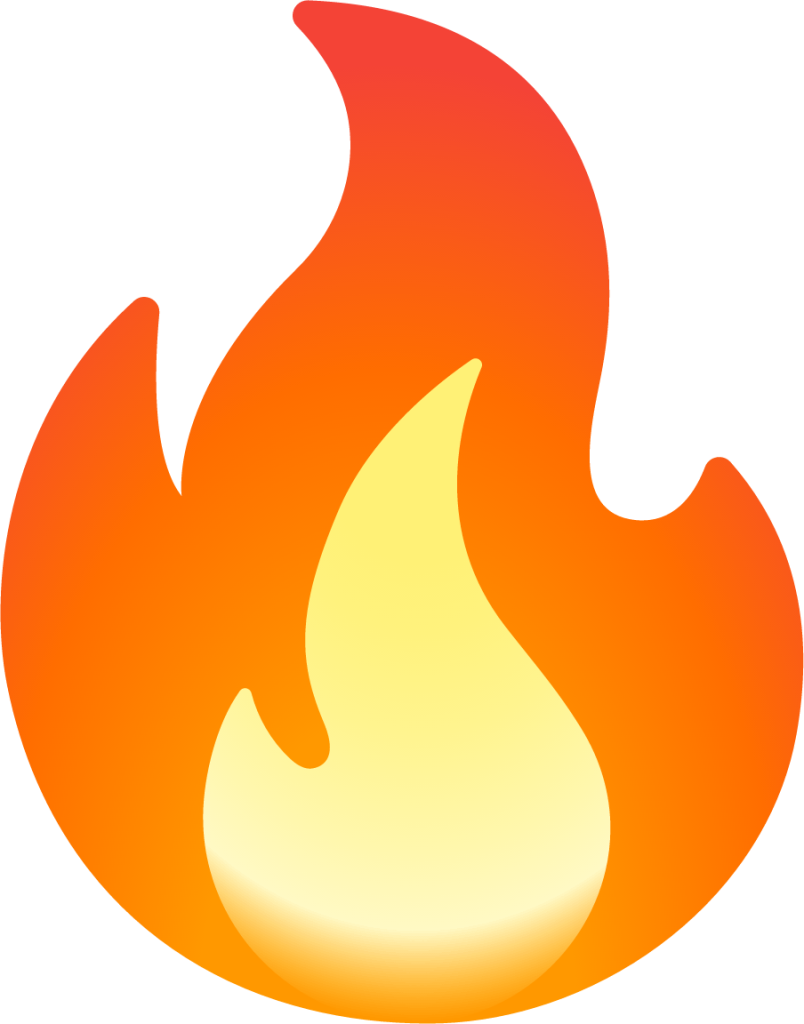}}~\textsc{Prometheus 2}: An Open Source Language Model Specialized in Evaluating Other Language Models}

\author{Seungone Kim$^{1,2,3}$\thanks{~~equal contribution. Work was done while Seungone was an intern at LG AI Research and a MS student at KAIST.} \quad Juyoung Suk$^{1}$$\textbf{}^{*}$ \quad Shayne Longpre$^{4}$ \quad \textbf{Bill Yuchen Lin}$^{5}$ \quad \textbf{Jamin Shin}$^{1}$\\ \textbf{Sean Welleck}$^{3}$ \quad \textbf{Graham Neubig}$^{3}$ \quad \textbf{Moontae Lee}$^{2,6}$ \quad \textbf{Kyungjae Lee}$^{2}$ \quad \textbf{Minjoon Seo}$^{1}$ \\ \\ KAIST AI$^{1}$ \qquad LG AI Research$^{2}$ \qquad Carnegie Mellon University$^{3}$ \qquad MIT$^{4}$\\ Allen Institute for AI$^{5}$ \qquad University of Illinois Chicago$^{6}$\\
\texttt{seungone@cmu.edu \qquad \{juyoung, minjoon\}@kaist.ac.kr}}

\begin{document}
\maketitle
\newcommand{\mcal}[1]{{\cal{#1}}}
\newcommand{\calA}{\mbox{${\cal A}$}}
\newcommand{\calB}{\mbox{${\cal B}$}}
\newcommand{\calC}{\mbox{${\cal C}$}}
\newcommand{\calD}{\mbox{${\cal D}$}}
\newcommand{\calE}{\mbox{${\cal E}$}}
\newcommand{\calF}{\mbox{${\cal F}$}}
\newcommand{\calG}{\mbox{${\cal G}$}}
\newcommand{\calH}{\mbox{${\cal H}$}}
\newcommand{\calI}{\mbox{${\cal I}$}}
\newcommand{\calJ}{\mbox{${\cal J}$}}
\newcommand{\calK}{\mbox{${\cal K}$}}
\newcommand{\calL}{\mbox{${\cal L}$}}
\newcommand{\calM}{\mbox{${\cal M}$}}
\newcommand{\calN}{\mbox{${\cal N}$}}
\newcommand{\calO}{\mbox{${\cal O}$}}
\newcommand{\calP}{\mbox{${\cal P}$}}
\newcommand{\calQ}{\mbox{${\cal Q}$}}
\newcommand{\calR}{\mbox{${\cal R}$}}
\newcommand{\calS}{\mbox{${\cal S}$}}
\newcommand{\calT}{\mbox{${\cal T}$}}
\newcommand{\calU}{\mbox{${\cal U}$}}
\newcommand{\calV}{\mbox{${\cal V}$}}
\newcommand{\calW}{\mbox{${\cal W}$}}
\newcommand{\calX}{\mbox{${\cal X}$}}
\newcommand{\calY}{\mbox{${\cal Y}$}}
\newcommand{\calZ}{\mbox{${\cal Z}$}}

\newcommand*\concat{\mathbin{\|}}

\newcommand{\che}[1]{\textcolor{brown}{#1}}

\newcommand{\se}{{\it SE}}
\newcommand{\eg}{{\it e.g.}}
\newcommand{\ie}{{\it i.e.}}
\newcommand{\etal}{{\it et al.}}
\newcommand{\etc}{{\it etc}}
\newcommand{\yeoo}{\textcolor{blue}}
\newcommand{\argmin}{\mathop{\mathrm{argmin}}\limits}
\newcommand{\argmax}{\mathop{\mathrm{argmax}}\limits}
\newcommand{\gc}{\textcolor{green}{\ding{52}}}
\newcommand{\bt}{\textcolor{black}{\ding{115}}}
\newcommand{\rx}{\textcolor{red}{\ding{55}}}
\definecolor{lightblue}{RGB}{224,236,247}
\definecolor{deepblue}{RGB}{9,46,107}
\begin{abstract}
Proprietary LMs such as GPT-4 are often employed to assess the quality of responses from various LMs. However, concerns including transparency, controllability, and affordability strongly motivate the development of open-source LMs specialized in evaluations. On the other hand, existing open evaluator LMs exhibit critical shortcomings: 1) they issue scores that significantly diverge from those assigned by humans, and 2) they lack the flexibility to perform both direct assessment and pairwise ranking, the two most prevalent forms of assessment. Additionally, they often do not possess the ability to evaluate based on \emph{custom evaluation criteria}, focusing instead on general attributes like helpfulness and harmlessness. To address these issues, we introduce Prometheus 2. Prometheus 2 is more powerful than its predecessor, and closely mirrors human and GPT-4 judgements. Moreover, it is capable of processing both direct assessment and pair-wise ranking formats grouped with a user-defined evaluation criteria. On four direct assessment benchmarks and four pairwise ranking benchmarks, \textsc{Prometheus 2} scores the highest correlation and agreement with humans and proprietary LM judges among all tested open evaluator LMs. Our models, code, and data are all publicly available.~\footnote{\href{https://github.com/prometheus-eval/prometheus-eval}{https://github.com/prometheus-eval/prometheus-eval}}
\end{abstract}

\begin{figure}[t!]
\centering
    \includegraphics[width=0.99\linewidth]{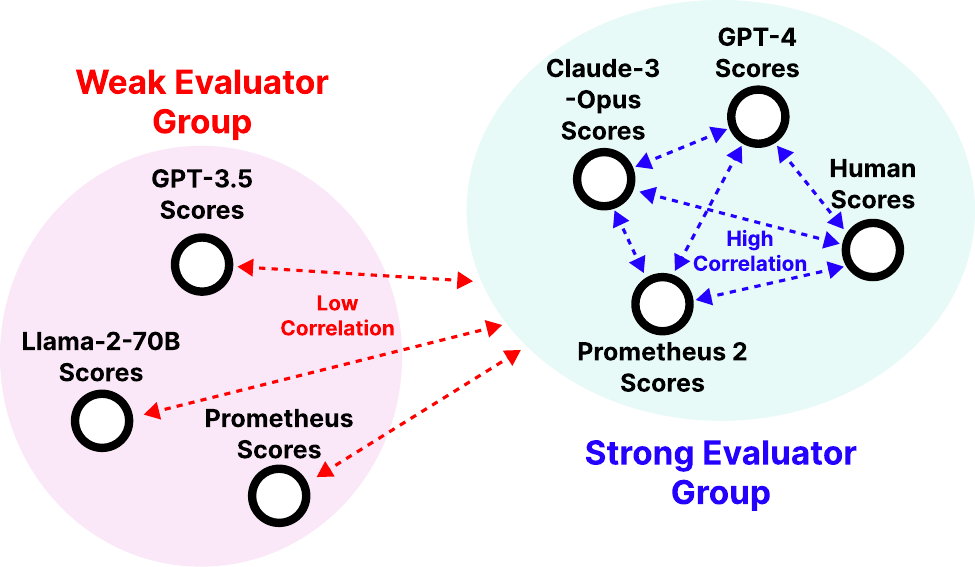}
    \caption{Weak evaluators (\textit{e.g.}, Llama-2-Chat-70B, Prometheus, and GPT-3.5-Turbo) achieve low scoring correlation with strong evaluators (\textit{e.g.}, Humans, GPT-4, and Claude-3-Opus). On the other hand, scores provided by strong evaluators highly correlate with each other.}
    \label{img:Promixtheus}
\vspace{-3mm}
\end{figure}

\section{Introduction}

Evaluating the quality of outputs produced by language models (LMs) is progressively becoming difficult, as the outputs cover an extremely diverse distribution of text and complex tasks. To address this issue, language model-based evaluation has emerged as a scalable and cheap paradigm for assessing LM-generated text~\citep{li2024leveraging,gao2024llm}. In this paradigm, LMs are either prompted to output a scalar indicator of quality (denoted as \textit{direct assessment})~\citep{zheng2023judging,liu2023gpteval,ye2023flask,kim2023prometheus} or to determine which of two outputs are preferred (denoted as \textit{pairwise ranking})~\citep{wang2023pandalm,alpaca_eval,lambert2024rewardbench}. Prior works employing proprietary LMs as evaluators have demonstrated not only high correlations with human evaluations but also increased speed and cost-effectiveness~\citep{zheng2023judging, liu2023gpteval,  dubois2023alpacafarm, ye2023flask}.

However, relying on proprietary LMs for evaluation poses significant challenges. The lack of transparency about their training data compromises both fairness and reproducibility, making it problematic to use them in evaluation pipelines. Additionally, concerns regarding controllability and affordability also persist~\citep{kim2023prometheus}. To address these issues, recent works have focused on developing evaluator LMs that are open-access, transparent, and controllable~\citep{kim2023prometheus, wang2023math, wang2023pandalm, li2023generative, zhu2023judgelm, jiang2023tigerscore, jiang2023llm, lee2024prometheus}. Yet, these models often yield scoring decisions that do not correlate well enough with human judgments or those made by proprietary LMs, failing to effectively simulate them. Moreover, open evaluator LMs are not flexible since they are typically trained only to perform either direct assessment or pairwise ranking and assess based on general public preferences like helpfulness and harmlessness, limiting their ability to handle diverse real-life scenarios.



To close the gap with proprietary LMs, we investigate \textit{unifying} the two model-based evaluation paradigms - direct assessment and pairwise ranking - to train a robust unified evaluator LM. We propose a recipe based on merging the weights of two evaluator LMs trained separately on direct assessment and pairwise ranking formats. Our key empirical observation is that weight merging can yield an evaluator LM that not only \textit{works} in both formats, but also \textit{outperforms} evaluator LMs that are jointly trained or only trained on a single format.

To demonstrate our approach, we develop the \textsc{Preference Collection}, a new fine-grained pairwise ranking feedback dataset that builds on the \textsc{Feedback Collection}~\cite{kim2023prometheus}, which is a direct assessment feedback dataset. We choose Mistral-7B~\citep{jiang2023mistral} and Mixtral-8x7B~\citep{jiang2024mixtral} as our base models, and merge the weights of evaluator LMs separately trained on the \textsc{Feedback Collection} and the \textsc{Preference Collection} to obtain our resulting models, \textsc{Prometheus 2} (7B \& 8x7B).



On four direct assessment benchmarks (Vicuna Bench, MT Bench, FLASK, Feedback Bench), the \textsc{Prometheus 2} models demonstrate the highest correlation with both human evaluators and proprietary LM-based judges compared to existing open evaluator LMs, with the Pearson correlation surpassing other baselines by 0.2 units across all datasets. Similarly, on four pairwise ranking benchmarks (HHH Alignment, MT Bench Human Judgment, Auto-J Eval, Preference Bench), the \textsc{Prometheus 2} models show the highest agreement with human evaluators among all the open evaluator LMs we tested, reducing the performance gap with GPT-4 in half.


Our contributions are summarized as follows:
\begin{itemize}
    \item We introduce \textsc{Prometheus 2} (7B \& 8x7B), state-of-the-art open evaluator LMs that score high correlations with both human evaluators and proprietary LM-based judges on both direct assessment and pairwise ranking.
    \item We introduce a pairwise ranking feedback dataset called the \textsc{Preference Collection}, which includes 1K custom evaluation criteria beyond helpfulness and harmlessness.
    \item We show that merging the weights of evaluator LMs trained on direct assessment and pairwise ranking feedback datasets results in a unified evaluator LM that excels in both schemes.
\end{itemize}

\section{Related Work}

\subsection{Language Model-based Evaluation}

To assess the generation capabilities of LMs, prior works such as the GEM benchmark~\citep{gehrmann2021gem,gehrmann2022gemv2} employed ROUGE~\citep{lin2004rouge}, BLEU~\citep{papineni2002bleu}, and BERTScore~\citep{zhang2019bertscore} as their metrics, which measure the lexical or semantic similarity between a reference answer and a response. However, these conventional metrics are prone to false negatives because they are not expressive enough to recognize responses that are of good quality but differ from the reference answer~\citep{schluter2017limits, freitag2020bleu, hanna2021fine}.

Recently, employing language models as a judge has gained attention as a promising paradigm to mimic the depth and granularity that human evaluation offers~\citep{zheng2023judging, liu2023gpteval, alpaca_eval,  chan2023chateval, ye2023flask}. To reduce the over-reliance on proprietary LMs, follow-up works suggest training language models specialized in evaluations~\citep{cui2023ultrafeedback,kim2023prometheus,jiang2023tigerscore,jiang2023llm,li2023generative,lee2024prometheus}. Yet, open evaluator LMs do not possess the flexibility to function in different evaluation schemes and show weak evaluation performance compared to proprietary LMs. We aim to bridge this gap by introducing \textsc{Prometheus 2}.

\subsection{Weight Merging}

Prior works have demonstrated that weight merging can enhance performance across various domains, including language modeling~\citep{li2022branch, matena2022merging, ilharco2022editing, don2022cold, gururangan2023scaling, yadav2024ties, sukhbaatar2024branch}, instruction-tuning~\citep{jang2023exploring,yu2023language}, and aligning to user preferences~\citep{jang2023personalized, rame2024rewarded, wang2024arithmetic}. In our work, we specifically focus on enhancing the evaluation capabilities of open evaluator LMs. By merging models trained on different assessment formats—specifically, direct assessment and pairwise ranking—we aim to obtain an evaluator LM that not only functions in both formats but also shows as good evaluation performances as proprietary LMs.

\begin{figure*}[t!]
\centering
    \includegraphics[width=0.99\linewidth]{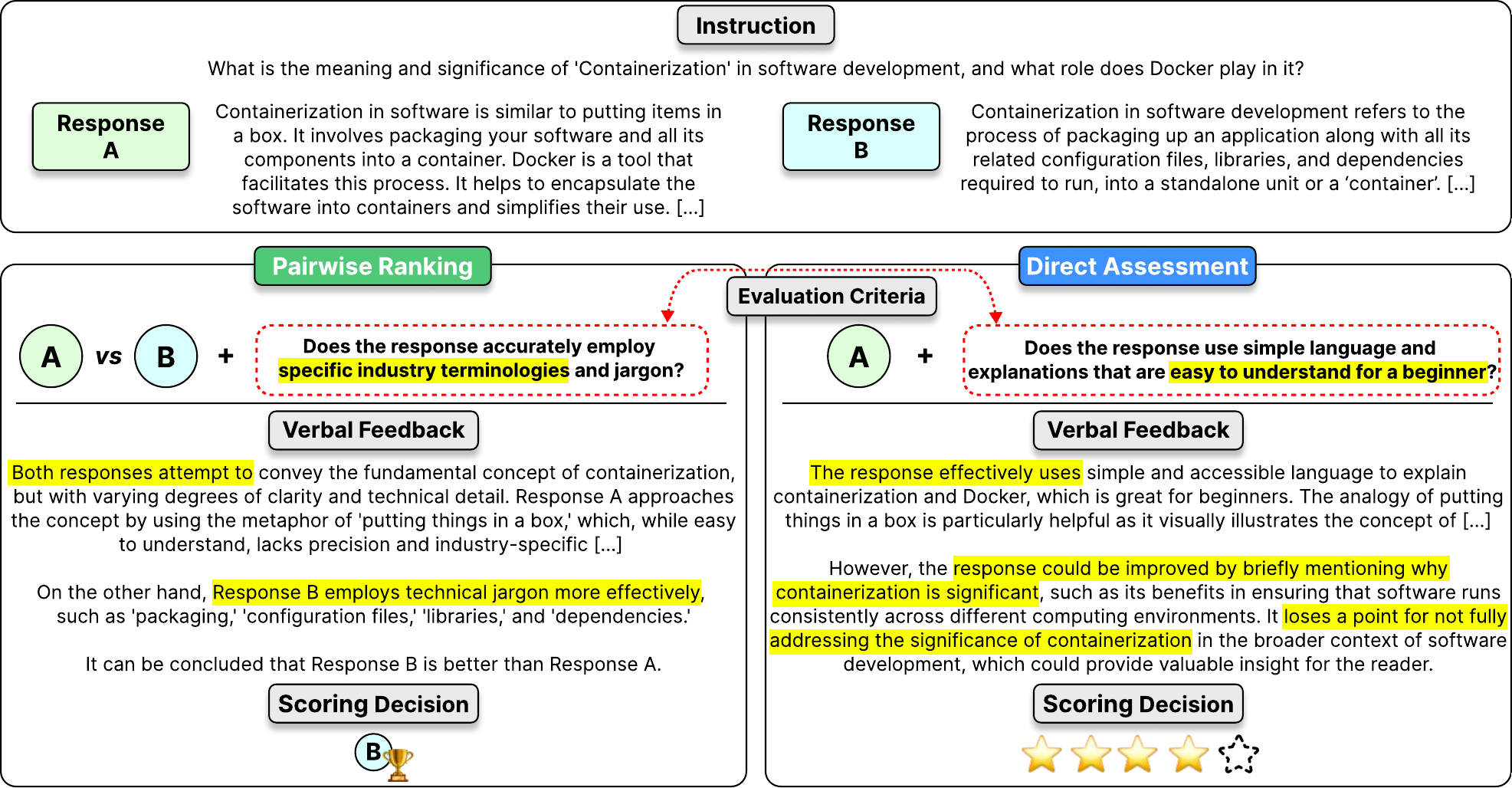}
    \caption{Comparison of direct assessment and pairwise ranking. Both responses could be considered decent under the umbrella of `helpfulness'. However, the scoring decision might change based on a specific evaluation criterion.}
    \label{img:Promixtheus}
\vspace{-3mm}
\end{figure*}

\section{Methodology}
We propose a new recipe for training a unified evaluator LM based on merging the weights of models trained for direct assessment and pairwise ranking. We begin with background on direct assessment and pairwise ranking for evaluator LMs (Section~\ref{critique_modeling}, \ref{reward_modeling}), followed by the construction process of our training data (Section~\ref{preference_collection}). Finally, we present our methods to train state-of-the-art evaluator LMs, Prometheus 2 models (Section~\ref{unified_training}).

\subsection{Direct Assessment}\label{critique_modeling}

Direct assessment is mapping an instruction $i$ and response $r$ into a scalar value score $s$, such as ${f}_{direct}: (i, r) \mapsto s \text{ where } s \in \mathbb{R}$. For the scoring range, we use an integer between 1 and 5. 

Prior works have identified several recipes to align the scores provided by evaluator LMs (${s}_{LM}$) and the scores assigned by humans (${s}_{human}$). For instance, \citet{liu2023geval} and \citet{zheng2023judging} have shown that it is crucial to add a reference answer $a$ as input to the evaluator LM to maximize the correlation between ${s}_{LM}$ and ${s}_{human}$. Also, \citet{zheng2023judging} and \citet{ye2023flask} showed that prompting the language model to write verbal feedback ${v}_{r}$ before $s$ also improves the correlation between ${s}_{LM}$ and ${s}_{human}$. Lastly, \citet{ye2023flask} and \citet{kim2023prometheus} showed that by explicitly integrating evaluation criteria $e$, users can define the standards for model assessment, ensuring evaluations are flexible to specific needs rather than generic qualities. Specifically, $e$ is represented as a score rubric including a description for the criterion itself and a set of descriptions for each score between the scoring range. This is expressed as: 
\begin{equation}
\begin{aligned}
f_{\text{direct}}: (i, r, a, e) &\mapsto ({v}_{r}, s) \\
\text{where } s &\in \{1, 2, 3, 4, 5\}
\end{aligned}
\end{equation}



\subsection{Pairwise Ranking}\label{reward_modeling}

Pairwise ranking is mapping an instruction $i$ and two pair of responses $({r}_{m}$, ${r}_{n})$ into either $i$ or $j$, such as ${f}_{pair}: (i, r_m, r_n) \mapsto s \text{ where } s \in \{m, n\}$.


Similar to direct assessment, prior works have identified that integrating a reference answer $a$ and verbal feedback ${v}_{{r}_{m},{r}_{n}}$ into the evaluation pipeline is crucial~\citep{zheng2023judging,alpaca_eval,li2023generative}. In addition, to support granular assessment under custom criterion, we add the evaluation criteria $e$ as input to the evaluator LM~\citep{ye2023flask,kim2023prometheus}. To the best of our knowledge, we are the first to study such fine-grained evaluation in pairwise ranking settings. This is expressed as: 
\begin{equation}
\begin{aligned}
    {f}_{pair}: (i, r_m, r_n, a, e) \mapsto ({v}_{{r}_{m},{r}_{n}}, s)\\ 
    \text{ where } s \in \{m, n\}   
\end{aligned}
\end{equation}

In pairwise ranking, the evaluation criterion $e$ does not include a set of descriptions for each score; instead, only the description of the evaluation criterion itself. Also, it is noteworthy that the verbal feedback ${v}_{{r}_{m},{r}_{n}}$ compares the commonalities and differences between ${r}_{m}$ and ${r}_{n}$ concerning $e$.


\begin{table}[t]
\centering
\fontsize{7.5}{10}\selectfont
{
\begin{tabular}{ccc}
\toprule
\multirow{2}{*}{\textbf{Data}} & \textsc{Preference} & \textsc{Feedback}\\
&\textsc{Collection} & \textsc{Collection}\\
\midrule
\textbf{Evaluation Scheme} & Pairwise Ranking & Direct Assessment\\
\textbf{\# Evaluation Criteria} & 1,000 & 1,000\\
\textbf{\# Instructions} & 20,000 & 20,000 \\
\textbf{\# Reference Answer} & 20,000 & 20,000\\
\textbf{\# Instances} & 200,000 & 100,000\\
\textbf{\#Verbal Feedback} & 200,000 & 100,000\\
\bottomrule
\end{tabular}%
}

\caption{Statistics of our training datasets, the \textsc{Feedback Collection} and the \textsc{Preference Collection}. Note that the 1K evaluation criteria, 20K instructions, and 20K reference answers are \textit{shared } among the two datasets. Both datasets have an equal number of scoring decisions (``A'' or ``B''; 100K each \& 1-5; 20K each) to prevent unintended biases after training.}
\label{tab:train-sets}
\end{table}

\subsection{The Preference Collection}\label{preference_collection}

Popular pairwise ranking datasets such as HH-RLHF~\citep{bai2022training} or Ultra Feedback~\citep{cui2023ultrafeedback} do not include an evaluation criterion $e$ and a verbal feedback ${v}_{{r}_{m},{r}_{n}}$. To train an evaluator LM that could assess based on such criteria, we construct the \textsc{Preference Collection}, including 1K evaluation criteria. We apply two modifications to the \textsc{Feedback Collection}. First, since the \textsc{Feedback Collection} includes five responses for each instruction, each corresponding to a scoring decision between 1 and 5, we pair two out of the five responses, resulting in a total of ten combinations per instruction. Using the existing scoring decisions for each response, we determine which response is better and assign a new scoring decision for that pair (\textit{i.e.}, ``Response A is better'' or ``Response B is better''). Second, to generate new verbal feedback ${v}_{{r}_{m},{r}_{n}}$ for each pair of responses, we prompt GPT-4-1106 to identify the commonalities and differences between the two responses. 

The statistics of the resulting dataset are listed in Table~\ref{tab:train-sets} along with the \textsc{Feedback Collection}. We explain about our quality verification process of the \textsc{Preference Collection} in Appendix~\ref{appendix:quality_verification}. Also, we include the prompts we use for the augmentation process in Appendix~\ref{appendix:augmentation_prompt}.

\begin{table*}[t]
    \centering
    \fontsize{6}{8}\selectfont{
        \begin{tabular}{@{}ccccccccccc@{}}
            \toprule
            \textbf{Evaluation Method}& \textbf{Benchmark} & \textbf{Metrics} & \textbf{Judgment Source} & \textbf{Reference Answer} & \textbf{\# Score Rubrics} & \textbf{\# Instructions} & \textbf{\# Judgments} \\
            \midrule
             \multicolumn{1}{l}{\multirow{4}{*}{\textbf{Direct Assessment}}} &\textbf{Vicuna Bench} & Correlation & Proprietary LMs & Y & 80 & 80 & 320 \\
            &\textbf{MT Bench} & Correlation & Proprietary LMs & Y & 80 & 80 & 320 \\
            &\textbf{FLASK} & Correlation & Proprietary LMs \& Humans & Y & 12 & 200 & 2,000 \\
            &\textbf{Feedback Bench} & Correlation & Proprietary LMs & Y & 200 & 200 & 1,000 \\
            \midrule
            \multicolumn{1}{l}{\multirow{4}{*}{\textbf{Pairwise Ranking}}} & \textbf{HHH Align.} & Accuracy & Humans & N & 4 & 221 & 221 \\
            &\textbf{MT Bench Human Judg.} & Accuracy & Humans & N & 1 & 80 & 3,360 \\
            &\textbf{Auto-J Eval} & Accuracy & Humans & N & 1 & 58 & 1,392 \\
            &\textbf{Preference Bench} & Accuracy & Proprietary LMs & Y & 200 & 200 & 2,000 \\
            \bottomrule
        \end{tabular}
    }
    \caption{Statistics of our evaluation benchmarks to assess the evaluation capabilities of evaluator LMs.}
    \label{tab:eval-sets}
\end{table*}

\subsection{Training Methods \& Baselines}\label{unified_training}

\paragraph{Prompting} Prompting involves querying an LM to make judgments in a specified evaluation format without training. We employ Llama-2-Chat-{7,13,70}B~\citep{touvron2023llama2}; Mistral-7B-Instruct-v0.2~\citep{jiang2023mistral}; and Mixtral-8x7B-Instruct-v0.1~\citep{jiang2024mixtral} as our baselines. It's worth noting that models not explicitly trained on feedback data often fail to generate responses in the required format, making it extremely difficult to parse scoring decisions. Although it is impractical for regular use, we make a fair comparison by infinitely looping until scores can be parsed. Also, we include proprietary LMs such as GPT-3.5-Turbo-0613; GPT-4-1106; and Claude-3-Opus.

\paragraph{Single-Format Training} Single-Format training involves training a base model $\theta$ on either on a direct assessment feedback dataset ${D}_{d}$ or a pairwise ranking feedback dataset ${D}_{p}$. For single-format trained evaluator LMs, we test Prometheus-{7,13}B~\citep{kim2023prometheus} (direct assessment); UltraRM-13B~\citep{cui2023ultrafeedback} (pairwise ranking); and PairRM-0.4B~\citep{jiang2023llm} (pairwise ranking). In addition, we also report the performances of single-format training Mistral-7B-Instruct-v0.2 and Mixtral-8x7B-Instruct-v0.1 on either direct assessment or pairwise ranking.

\paragraph{Joint Training} Joint training involves training a base model $\theta$ on both a direct assessment feedback dataset ${D}_{d}$ and a pairwise ranking feedback dataset ${D}_{p}$. This enables the resulting evaluator LM to function across both evaluation formats. For jointly trained evaluator LMs, we test Auto-J~\citep{li2023generative}. In addition, we report the performances of jointly training Mistral-7B and Mixtral-8x7B on both direct assessment and pairwise ranking.

\paragraph{Weight Merging} Weight Merging involves training two models, ${\theta}_{d}$ and ${\theta}_{p}$, separately on a direct assessment feedback dataset ${D}_{d}$ and a pairwise ranking feedback dataset ${D}_{p}$. Then, the final evaluator LM ${\theta}_{final}$ is obtained by merging ${\theta}_{d}$ and ${\theta}_{p}$. For example, linear merging is as follows:


\begin{equation}
    {\theta}_{final} = \alpha \times {\theta}_{d} + (1 - \alpha) \times {\theta}_{p}
\end{equation}

In addition to linear merging, we test 5 additional variants, namely Task Arithmetic merging~\citep{ilharco2022editing}, TIES merging~\citep{yadav2024ties}, DARE-TIES and DARE-Linear merging~\citep{yu2023language}, and SLERP merging~\citep{goddard2024arcee}. We include an explanation of these merging methods and ablation experiment results of the performance differences in Appendix~\ref{appendix:merging}. Among them, DARE-Linear showed the best performance, and hence we used it to train the \textsc{Prometheus 2} (7B \& 8x7B) models. Details on the hyper-parameters for training and inference along with the prompt templates are all listed in Appendix~\ref{appendix:hyperparameter}, \ref{appendix:direct_assessment_prompt}, \ref{appendix:pairwise_ranking_prompt}.



\section{Experimental Setup}\label{experimental_setting}


The statistics of all the benchmarks are in Table~\ref{tab:eval-sets}. The four direct assessment benchmarks are:

\begin{itemize}
    \item \textbf{Vicuna Bench}~\citep{vicuna2023}: A single-turn chat benchmark that includes 80 test prompts, 80 hand-crafted score rubrics from \citet{kim2023prometheus}, and 320 responses obtained by WizardLM-13B, Vicuna-13B, Llama-2-Chat-13B, GPT-3.5-Turbo-0613.
    \item \textbf{MT Bench}~\citep{zheng2023judging}: A multi-turn chat benchmark that consists of 80 test prompts, 80 hand-crafted score rubrics from \citet{kim2023prometheus}, and 320 responses obtained by WizardLM-13B, Vicuna-13B, Llama-2-Chat-13B, GPT-3.5-Turbo-0613.
    \item \textbf{FLASK}~\citep{ye2023flask}: A fine-grained evaluation benchmark comprised of 200 test prompts, 12 score rubrics, and 2000 responses acquired from Alpaca-7B, Vicuna-13B, Bard, GPT-3.5-Turbo-0613. In addition to scores from proprietary LMs, this benchmark also includes scores marked by human evaluators.
    \item \textbf{Feedback Bench}~\citep{kim2023prometheus}: The test set of the \textsc{Feedback Collection} with 1K score rubrics, 200 instructions, and 1K responses that do not overlap with the train data.
\end{itemize}

\begin{table*}[t!]
\fontsize{9}{14}\selectfont
\centering
\resizebox{\textwidth}{!}{\begin{tabular}{lcccccccc}
    \toprule
    \multicolumn{1}{c}{\multirow{2}{*}{\textbf{Evaluator LM}}}& \multicolumn{2}{c}{\textsc{Vicuna Bench}} & \multicolumn{2}{c}{\textsc{MT Bench}} & \multicolumn{3}{c}{\textsc{FLASK}} & Feedback Bench\\ 
    \cmidrule(lr){2-3} \cmidrule(lr){4-5} \cmidrule(lr){6-8} \cmidrule(lr){9-9} & GPT-4-1106& Claude-3-Opus & GPT-4-1106& Claude-3-Opus & GPT-4-1106& Claude-3-Opus & Humans & GPT-4-0613\\
    \midrule
    \textsc{Llama2-Chat 7B}&0.205&0.243&0.036&0.055&0.317& 0.256& 0.299&0.523\\
    \textsc{Llama2-Chat 13B}&0.185 &0.141 & -0.042 & -0.002 & 0.239 & 0.247 & 0.263&0.545\\
    \textsc{Llama2-Chat 70B}&0.350&0.463&0.178&0.228&0.388& 0.402 & 0.317&0.592\\
    \textsc{Mistral-Instruct-7B}&0.486&0.561&0.284&0.396&0.448& 0.437 & 0.377&0.586\\
    \textsc{Mixtral-Instruct-8x7B}&0.566&0.579&0.551&0.539&0.483& 0.495 & 0.420&0.673\\
    \textsc{Prometheus-7B}&0.484&0.528&0.378&0.382&0.352&0.331 & 0.348&0.847\\
    \textsc{Prometheus-13B}&0.492&0.534&0.404&0.477&0.462&0.470 & 0.449&0.860\\
    \textsc{Auto-J (13B)}&0.351&0.262&0.432&0.375&0.430& 0.370 & 0.473&0.637\\
    \textsc{Prometheus-2-7B}&\underline{0.666}&\textbf{0.654}&\underline{0.548}&\underline{0.517}&\underline{0.617}& \underline{0.561}& \underline{0.545}&\underline{0.882}\\
    \textsc{Prometheus-2-8x7B}&\textbf{0.685}&\underline{0.635}&\textbf{0.665}&\textbf{0.614}&\textbf{0.659}& \textbf{0.626} & \textbf{0.555}&\textbf{0.898}\\
    \midrule
    \textsc{GPT-3.5-Turbo-0613}&0.335&0.349&0.183&0.194&0.437& 0.396 & 0.450&0.594\\
    \textsc{GPT-4-1106}&/&0.694&/&0.717&/& 0.736 & 0.679&0.753\\
    \textsc{Claude-3-Opus}&0.694&/&0.717&/&0.736&/&0.573&0.788\\
    \bottomrule    
\end{tabular}}
\caption{\footnotesize \textbf{Direct Assessment Results} Pearson correlations between reference evaluators (listed on top) and evaluator LMs. The best comparable statistics are \textbf{bolded} and second best \underline{underlined} except proprietary LMs. Spearman and Kendall-Tau correlations are reported in Appendix~\ref{appendix:direct-assessment}. Note that the Feedback Bench is an in-domain test set of the \textsc{Prometheus} models.}
\label{table:direct-assessment}
\end{table*}

The four pairwise ranking benchmarks are:

\begin{itemize}
    \item \textbf{HHH Alignment}~\citep{askell2021general}: A benchmark consisting of 221 prompts; 4 score rubrics (helpfulness, harmlessness, honesty, and other) and 221 response pairs (graded as `win' or `lose') judged by human evaluators.
    \item \textbf{MT Bench Human Judgment}~\citep{zheng2023judging}: A benchmark that shares the same 80 prompts as MT-Bench. In addition, it provides 3,360 response pairs (graded as `win', `tie', or `lose') judged by human evaluators.
    \item \textbf{Auto-J Eval}~\citep{li2023generative}: A benchmark consisted of 58 prompts and 1,392 response pairs (graded as `win', `tie', or `lose') judged by human evaluators. This benchmark is used as the in-domain test set of Auto-J.
    \item \textbf{Preference Bench}: Our in-domain test set for the \textsc{Prometheus} models. Similar to how the \textsc{Preference Collection} was made with the \textsc{Feedback Collection}, we adjust the \textsc{Feedback Bench} and pair two out of the five responses, resulting in a test set with 200 prompts, 2,000 response pairs (graded as `win' or `lose'), and 200 evaluation criteria.
\end{itemize}

In direct assessment, we conduct \textbf{reference-based} evaluations by appending the reference answer as the input. We use \textbf{Pearson}, \textbf{Spearman}, and \textbf{Kendall-Tau} as performance metrics to measure scoring correlations against reference evaluators. Moreover, we include the results of the reference-free direct assessment evaluation in Appendix~\ref{appendix:reference_free}.

In pairwise ranking, we conduct \textbf{reference-free} evaluations. Based on judgments assigned by humans, we use \textbf{accuracy} as our metric to measure agreement between evaluator LMs and humans.

Also, the MT Bench Human Judgment and Auto-J test set includes a `tie' option assessed by human evaluators. We evaluate in two ways: by excluding all `tie' options for pairwise ranking (denoted as `\textbf{w/o tie}'), or by using direct assessment where responses scored as `ties' are grouped, and pairwise rankings are applied to the remaining responses with differing scores (denoted as `\textbf{w/ tie}').

\begin{table*}[t!]
\centering
\fontsize{6}{8}\selectfont
\resizebox{\textwidth}{!}{\begin{tabular}{lccccccccccc}
    \toprule
    \multicolumn{1}{c}{\multirow{2}{*}{\textbf{Evaluator LM}}}& \multicolumn{5}{c}{\textsc{HHH Alignment}} & \multicolumn{2}{c}{\textsc{MT Bench Human Judg.}} & \multicolumn{2}{c}{\textsc{Auto-J Eval}} & Preference Bench\\ 
    \cmidrule(lr){2-6} \cmidrule(lr){7-8} \cmidrule(lr){9-10} \cmidrule(lr){11-11} & Help. & Harm. & Hon. & Other & Total Avg. & w/ TIE & w/o TIE & w/ TIE & w/o TIE & Instance-wise Criteria\\ 
    \midrule
    \textsc{Llama2-Chat 7B}&55.93&62.07&49.18&62.79&57.01&46.68&50.39&45.76&45.73 & 58.60\\
    \textsc{Llama2-Chat 13B}&71.19&77.59&60.66&62.79&68.33&51.22&49.61&47.84&43.28 & 63.00\\
    \textsc{Llama2-Chat 70B}&62.71&81.03&65.57&65.12&68.78&\textbf{55.14}&60.88&53.38&50.64 & 64.70\\
    \textsc{Mistral-Instruct-7B}&59.32&68.97&63.93&81.40&67.42&53.81&63.82&53.88&60.94 & 79.40\\
    \textsc{Mixtral-Instruct-8x7B}&83.05&\underline{87.93}&67.21&69.77&77.38&51.85&\underline{71.42}&53.81&73.50 & 84.00\\
    \textsc{Pair RM (0.4B)}&\underline{84.75}&84.48&\underline{80.33}&\textbf{90.70}&\underline{84.62}&-&59.00&-&59.05 & 81.80\\
    \textsc{Ultra RM (13B)}&\textbf{86.44}&79.31&\textbf{81.97}&\underline{88.37}&83.71&-&56.00&-&59.85 & 86.97\\
    \textsc{Auto-J (13B)}&77.97&79.31&70.49&74.42&75.57& 42.56 & 69.12 & 43.46 & \underline{76.64} & 81.35\\
    \textsc{Prometheus-2-7B}&72.78& 79.31&77.05&76.74&74.66&50.45&70.78&\underline{54.96}&75.07 & \textbf{93.25}\\
    \textsc{Prometheus-2-8x7B}&\underline{84.75}&\textbf{96.55}&\textbf{81.97}&76.74&\textbf{85.52}&\underline{55.07}&\textbf{71.96}&\textbf{58.41}&\textbf{79.98} & \underline{90.65}\\
    \midrule
    \textsc{GPT-3.5-Turbo-0613}&77.97&81.03&77.05&67.44&76.47&54.65&69.41&45.98&72.13 & 75.05\\
    \textsc{GPT-4-1106-Preview}&89.83&96.55&91.80&83.72&90.95&60.38&79.90&52.80&83.12 & 85.50\\
    \textsc{Claude-3-Opus}&91.53&100.00&91.80& 95.35&94.57&55.35&77.65&60.70&82.92&89.85\\
    \bottomrule    
\end{tabular}}
\caption{\footnotesize \textbf{Pairwise Ranking Results} Accuracy on human preference datasets. The best comparable accuracies are \textbf{bolded} and second best \underline{underlined} except proprietary LMs. Note that HHH Alignment is an in-domain test set for PairRM, Auto-J Eval is an in-domain test set for Auto-J, and the Preference Bench is an in-domain test set for Prometheus-2 models.}
\label{table:pairwise-ranking}
\end{table*}

\section{Experimental Results}\label{experimental_results}

In this section, we compare the evaluation capabilities of \textsc{Prometheus-2} models with other baselines using a direct assessment format (Section~\ref{sec:direct_assessment}) and a pairwise ranking format (Section~\ref{sec:pairwise_ranking}). Additionally, we measure the consistency of the scores from evaluator LMs in Appendix~\ref{appendix:consistency}.

\subsection{Direct Assessment Results}\label{sec:direct_assessment} 
The direct assessment results are shown in Table~\ref{table:direct-assessment}. The scoring decisions of \textsc{Prometheus 2} models (7B \& 8x7B), GPT-4-1106, Claude-3-Opus, and human evaluators all strongly correlate with each other, yielding Pearson correlations higher than 0.5 regardless of the reference evaluator and benchmark. On the other hand, base LMs, single-format trained LMs, and jointly trained LMs show lower correlations, mostly falling below 0.5.

Notably, \textsc{Prometheus 2} models outperform Prometheus and Auto-J by at least 0.2 units across benchmarks in their correlation with proprietary LMs. Moreover, on the FLASK benchmark, while the correlation between humans and GPT-4 is 0.679, the highest correlation previously achieved by Prometheus-13B with humans was 0.449. \textsc{Prometheus-2-8x7B} achieves a correlation of 0.555 with humans, halving the gap.

\subsection{Pairwise Ranking Results}\label{sec:pairwise_ranking}
The pairwise ranking results are shown in Table~\ref{table:pairwise-ranking}. We exclude the results of Pair RM and Ultra RM on `w/ Tie' settings since they could not process it. 

On all of the 4 benchmarks, the \textsc{Prometheus 2} models achieve the highest scores, showing that they could effectively simulate human judgments. Notably, while HHH Alignment is an in-domain test set for Pair RM, and Auto-J Eval is for Auto-J, \textsc{Prometheus-2-8x7B} achieves higher scores. This shows that training a large LM (\textit{i.e.}, Mixtral-8x7B) with feedback data could be an effective strategy to obtain a robust evaluator LM that could generalize beyond its training data. Moreover, the \textsc{Prometheus 2} models at least halve the performance gap with proprietary LMs compared to existing evaluator LMs on out-of-domain test sets.

\begin{table*}[t!]
\centering
\fontsize{7}{10}\selectfont
\begin{tabular}{lcccccccc}
    \toprule
     \multicolumn{1}{c}{\multirow{2}{*}{\textbf{Training Method}}}&\multicolumn{4}{c}{\textsc{Direct Assessment Benchmarks}} & \multicolumn{4}{c}{\textsc{Pairwise Ranking Benchmarks}}\\ 
    \cmidrule(lr){2-5} \cmidrule(lr){6-9} & Vicuna Ben. & MT Ben. & FLASK & Average & HHH Align. & MT Ben. H.J. & Auto-J Eval & Average\\
    \midrule
    \textit{\textbf{Mistral-Instruct-7B}}\\
    \midrule
    \textsc{Prompting} & 0.486 & 0.284 & 0.480 & 0.417 &67.42&63.82&60.94&64.06\\
    \textsc{Direct Assessment Only} & 0.537 & \textbf{0.561} & \underline{0.519}  & \underline{0.539} & 73.33 & 56.76 & 64.38&64.82\\
    \textsc{Pairwise Ranking Only} & - & - & - & - & \underline{78.73} & \underline{67.06} & 72.03&72.61\\
    \textsc{Joint Training} & \underline{0.548} & 0.450 & 0.457 & 0.485 & \textbf{80.09} & 65.49 & \underline{73.60}&\underline{73.06}\\
    \textsc{Weight Merging} & \textbf{0.666} & \underline{0.548} & \textbf{0.659} & \textbf{0.624} & 74.66 & \textbf{70.78} & \textbf{75.07}&\textbf{73.50} \\
    \midrule
    \textit{\textbf{Mixtral-Instruct-8x7B}}\\
    \midrule
    \textsc{Prompting} & 0.566 & 0.551 & 0.507 & 0.541 & 77.38 & \underline{71.42} & 73.55 & 74.56\\
    \textsc{Direct Assessment Only} & 0.625 & \underline{0.664} & 0.587 & \underline{0.625} & 74.21 & 53.14 & 65.85 & 64.40\\
    \textsc{Pairwise Ranking Only} & - & - & - & - & \underline{84.16} & 66.27 & \underline{75.66}&\underline{75.36}\\
    \textsc{Joint Training} & \underline{0.628} & 0.560 & \underline{0.596} & 0.595 & 82.35 & 68.73 & 74.78&75.29\\
    \textsc{Weight Merging} & \textbf{0.685} & \textbf{0.665} & \textbf{0.659} & \textbf{0.670}& \textbf{85.52} & \textbf{71.96} & \textbf{79.98} &\textbf{79.15}\\
    \bottomrule    
\end{tabular}
\caption{\footnotesize \textbf{Single-Format Training vs Joint Training vs Weight Merging} Pearson correlations between evaluator LMs trained with different methods and GPT-4-1106. Evaluator LMs trained with weight merging outperform single-format-trained and jointly-trained evaluator LMs across multiple benchmarks.}
\label{table:merging_multitask}
\end{table*}

\begin{table*}[t!]
\centering
\fontsize{6.5}{9}\selectfont
\begin{tabular}{lcccccccc}
    \toprule
     \multicolumn{1}{c}{\multirow{2}{*}{\textbf{Training Data Evaluation Format}}}&\multicolumn{4}{c}{\textsc{Direct Assessment Benchmarks}} & \multicolumn{4}{c}{\textsc{Pairwise Ranking Benchmarks}}\\ 
    \cmidrule(lr){2-5} \cmidrule(lr){6-9} &Vicuna Ben. & MT Ben. & FLASK & Average & HHH Align. & MT Ben. H.J. & Auto-J Eval & Average\\
    \midrule
    \textsc{No Training (Prompting)} & 0.486 & 0.284 & 0.480 & 0.417 &67.42&63.82&60.94&64.06\\
    \midrule
    \textsc{Direct Assessment Only} & 0.537 & \textbf{0.561} & \underline{0.519}  & \underline{0.539} & 73.33 & 56.76 & 64.38 & 64.82\\
    \textsc{Pairwise Ranking Only} &  - & - & - & - & \textbf{78.73} & \underline{67.06} & 72.03&\underline{72.61}\\
    \midrule
    \textsc{Direct Assessment \& Direct Assessment} & \underline{0.552} &0.493 &0.505 & 0.517 & 73.30 & 55.00 &63.69 &64.13\\
    \textsc{Pairwise Ranking \& Pairwise Ranking}&-&-&-&-&\underline{78.70}&65.20&\underline{72.72}&72.21\\
    \midrule
    \textsc{Direct Assessment \& Pairwise Ranking}& \textbf{0.666} & \underline{0.548} & \textbf{0.659} & \textbf{0.624} & 74.66 & \textbf{70.78} & \textbf{75.07}&\textbf{73.50} \\
    \bottomrule    
\end{tabular}
\caption{\footnotesize \textbf{Unifying Formats vs Ensembling} Pearson correlations with GPT-4-1106 (Vicuna Bench, MT Bench, FLASK) and agreement with human evaluators (HHH Alignment, MT Bench Human Judgment, Auto-J Eval). Merging models trained with the same evaluation formats (ensembling) underperforms merging models trained with different formats (unifying formats).}
\label{table:merging_ablations}
\end{table*}

\section{Analyses of Weight Merging}\label{sec:weight_merging_analysis}

To understand the effectiveness of our proposed weight merging method in the context of evaluations, we address the following research questions:
\begin{itemize}
    \item \textbf{RQ1}: Is weight merging more effective compared to joint training? (Section~\ref{sec:6.1})
    \item \textbf{RQ2}: Is the effectiveness of weight merging due to model ensembling? (Section~\ref{sec:6.2})
    \item \textbf{RQ3}: To what extent does learning with direct assessment help pairwise ranking performance, and vice versa? (Section~\ref{sec:6.3})
\end{itemize}

\subsection{Weight Merging vs Joint Training}\label{sec:6.1}

Table~\ref{table:merging_multitask} compares the performance of evaluator LMs trained via weight merging and joint training. Alongside this, we also add and compare the results of prompting and single-format training.

Surprisingly, evaluator LMs trained via joint training often show lower performance compared to those trained only in single-format, which indicates \textit{negative task transfer}. Specifically, evaluator LMs trained only on direct assessment formats obtain higher correlations compared to their jointly trained counterparts across different model scales. Similarly, evaluator LMs trained solely on pairwise ranking formats achieve higher average accuracy compared to those trained on multiple tasks, particularly when using Mixtral-8x7B as the base model.

On the other hand, evaluator LMs trained via weight merging show superior performance not only compared to jointly trained evaluator LMs but also single-format trained evaluator LMs, indicating \textit{positive task transfer}. Also, while both benefit each other, merging the pairwise ranking evaluator LM weights improves direct assessment performance more significantly than the reverse.

\begin{figure*}[t!]
\centering
    \includegraphics[width=0.63\linewidth]{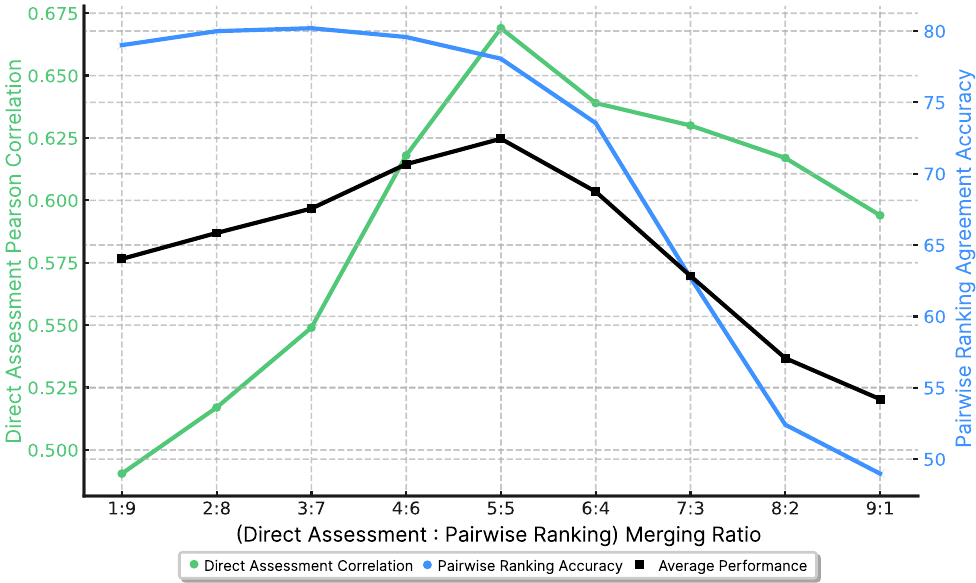}
    \caption{\textbf{When merging models, the influence of relative evaluation on absolute evaluation is greater than the influence of absolute evaluation on relative evaluation.} Performance of Direct Assessment (colored in green) and Pairwise Ranking (colored in blue) when altering the $\alpha$ value to merge evaluator LMs trained on different formats.}
    \label{img:alpha_optimal}
    \vspace{-3mm}
\end{figure*}

\subsection{Is the Effectiveness of Weight Merging due to Model Ensembling?}\label{sec:6.2}

While we empirically find that weight merging is effective, the underlying reason remains unclear. A natural assumption is that this effectiveness results from the ensembling effect of combining multiple models. To test this hypothesis, we conduct an ablation experiment where we train multiple evaluator LMs on different random seeds and merge them. Specifically, we merge two evaluator LMs trained on direct assessment formats (denoted as `Direct Assessment \& Direct Assessment') and two evaluator LMs trained on pairwise ranking formats (denoted as `Pairwise Ranking \& Pairwise Ranking'). We use Mistral-7B-Instruct as our base model.

The results are presented in Table~\ref{table:merging_ablations}. Across multiple benchmarks, merging evaluator LMs trained on the same evaluation format does not enhance evaluation performance. Specifically, merging two evaluator LMs trained on the same evaluation format—whether direct assessment or pairwise ranking—negatively impacts performance on average for both direct assessment and pairwise ranking benchmarks. In contrast, merging two evaluator LMs, each trained on direct assessment and pairwise ranking formats, results in superior performance compared to the other settings. This indicates that the beneficial task transfer in weight merging arises from integrating different evaluation formats, not ensembling multiple models.

\subsection{Quantifying Positive Transfer across Evaluation Formats}\label{sec:6.3}
To explore how training on direct assessment feedback data influences pairwise ranking accuracy and vice versa, we experiment by adjusting the $\alpha$ value during linear merging. We evaluate the average performance using all eight benchmarks in our experiments. To illustrate the average performance (colored in black), we adjust the scale by multiplying the Pearson correlations from direct assessment, which originally range from 0 to 1, by 100 before averaging them with the pairwise ranking accuracy.

The results are shown in Figure~\ref{img:alpha_optimal}. For direct assessment benchmarks, evaluator LMs obtain the optimal performance when $\alpha$ is set to 0.5. This indirectly indicates that both pairwise ranking and direct assessment feedback data contribute equally. On the other hand, for pairwise ranking benchmarks, the performance is optimal when $\alpha$ is set to 0.3. This also implies that while both benefit each other, training on pairwise ranking improves direct assessment performance more than the reverse.

\section{Conclusion}
We introduce \textsc{Prometheus 2}, an open-source LM specialized in evaluating other responses. Unlike existing open evaluator LMs that cannot effectively process both direct assessment and pairwise ranking—the two most prevalent evaluation schemes— the \textsc{Prometheus 2} models demonstrate superior performance on both schemes, significantly narrowing the gap with proprietary LM-based evaluations. To train the \textsc{Prometheus 2} models, we develop the \textsc{Preference Collection}, the first pairwise ranking dataset that includes over 1,000 instance-wise evaluation criteria beyond basic qualities such as helpfulness and harmlessness. Notably, we find that merging evaluator LMs trained on either direct assessment or pairwise ranking formats can lead to a unified evaluator LM with strong performance. We hope that our work encourages more research on using open-source LMs as evaluators.

\section*{Acknowledgements}
We thank the KAIST AI LKLab members for helpful discussions. This work was partly supported by LG AI Research grant (Self-improving logical reasoning capabilities of LLMs, 2024, 50\%) and the Institute of Information \& Communications Technology Planning \& Evaluation(IITP) grant funded by the Korea government(MSIT) (RS-2024-00397966, Development of a Cybersecurity Specialized RAG-based sLLM Model for Suppressing Gen-AI Malfunctions and Construction of a Publicly Demonstration Platform, 50\%).


\section*{Limitations}
Evaluation is fundamentally a very multi-faceted task. In this paper, we used an indirect method to assess the evaluation capability of evaluator LMs by measuring if they perform evaluations similar to human evaluators or proprietary LMs, such as GPT-4-1106 and Claude-3-Opus. However, this may not necessarily be the best approach. Future work could explore meta-evaluation pipelines that reevaluate the results of evaluator LMs or methodologies that allow humans to efficiently review evaluation results. Also note that it is crucial to use model-based evaluations in conjunction with human evaluation instead of solely relying on it.

Additionally, the degree to which evaluator LMs can generalize was based on an analysis by \citet{kim2023prometheus}, which checked for overlap between the data used to train the evaluator LMs and the data used to evaluate them. This study extended the evaluation to eight different datasets with human judgments to check the generalization capability of evaluation under various circumstances. However, this may not be sufficient. One of the major challenges in evaluating evaluator LMs is obtaining the ``evaluation results'' (\textit{e.g.}, human judgment). Automating evaluations with LMs could greatly benefit many areas of NLP research, hence the role of future work in creating feedback benchmarks that include human judgment or data for training evaluator LMs is crucial.

One downside of the \textsc{Prometheus 2} is that it operates only on a 1-5 point Likert scale for absolute evaluation or a comparative evaluation style of `A is better \& B is better'. Depending on the use case, people may need a 1-10 point absolute evaluation, a ranking method for five responses at once, or a checklist-based evaluation not covered in the paper. While proprietary LMs can flexibly conduct evaluations in any format if a well-described prompt is devised, open-source LMs cannot produce good evaluation results without training, and conversely, if trained in one or two formats, they lose the flexibility to conduct different evaluations. Future work could examine whether evaluator LMs trained in each format, as done in this paper, can handle evaluations for added formats well when weight merging is employed.

Lastly, the paper presents an evaluation model that can handle both absolute and comparative evaluation formats well through weight merging based on empirical experiments. However, fundamentally explaining why weight merging works well remains a challenging task. To address this, Section~\ref{sec:weight_merging_analysis} indirectly analyzes the effectiveness of weight merging by comparing it with joint training, demonstrating that the improvement in evaluation performance is not due to model ensembling, and showing that the impact of comparative evaluation on absolute evaluation is greater than the reverse. Our best current interpretation is that "absolute and comparative evaluations are not completely different tasks, so weight merging could handle both without degeneration, and conversely, because they are not too similar, weight merging performed better than joint training." Future work could theoretically analyze this or further explore whether weight merging can effectively work in fields other than LLM evaluation.







\bibliography{anthology,custom}
\bibliographystyle{acl_natbib}

\clearpage
\appendix

\begin{table}[t!]
\centering
\fontsize{8}{10}\selectfont
\begin{tabular}{cc}
\toprule
Verification Standards & \textsc{Results} \\ \midrule
\textbf{Coherence} & 99.5 \% (Passed)\\
\textbf{Suitability} &  98.5 \% (Passed)\\
\textbf{Criticality} & 88\% (Win rate) \\ 
\bottomrule
\end{tabular}%
\caption{\footnotesize Human verification results to assess the quality of the \textsc{Preference Collection}. We use three standards to assess the quality of verbal feedback ${v}_{{r}_{m},{r}_{n}}$.}
\label{tab:quality}
\end{table}

\begin{table}[t!]
\centering
\fontsize{8}{10}\selectfont
\begin{tabular}{c|c}
\toprule
\textbf{Temperature} & 1.0\\
\textbf{Top\_p} &  0.9\\
\textbf{Max New Tokens} & 1024 \\ 
\textbf{Repetition Penalty} & 1.03 \\ 
\bottomrule
\end{tabular}%
\caption{\footnotesize Hyperparameters used to inference different evaluator LM baselines.}
\label{tab:inference_hyperparameter}
\end{table}

\begin{table}[t!]
\centering
\fontsize{8}{10}\selectfont
\begin{tabular}{c|c}
\toprule
\textbf{Base Model} & mistralai/Mistral-7B-Instruct-v0.2\\
\textbf{Torch dtype} &  bfloat16\\
\textbf{Epoch} & 1 \\ 
\textbf{Train Data 1} & \textsc{Feedback Collection} \\ 
\textbf{Train Data 2} & \textsc{Preference Collection} \\ 
\textbf{Max Seq Length} & 4096\\
\textbf{Learning Rate} & 1e-5\\
\textbf{Train Batch Size} & 4\\
\textbf{Random Seed} & 42\\
\textbf{Merging Strategy} & Linear ($\alpha = 0.5$)\\
\textbf{Training Method} & Supervised Fine-tuning\\
\bottomrule
\end{tabular}%
\caption{\footnotesize Hyperparameters used to train \textsc{Prometheus 2} 7B.}
\label{tab:prometheus2_7b_hyperparameter}
\end{table}

\begin{table}[t!]
\centering
\fontsize{7}{9}\selectfont
\begin{tabular}{c|c}
\toprule
\textbf{Base Model} & mistralai/Mixtral-8x7B-Instruct-v0.1\\
\textbf{Torch dtype} &  bfloat16\\
\textbf{Epoch} & 1 \\ 
\textbf{Train Data 1} & \textsc{Feedback Collection} \\ 
\textbf{Train Data 2} & \textsc{Preference Collection} \\ 
\textbf{Max Seq Length} & 4096\\
\textbf{Learning Rate} & 1e-5\\
\textbf{Train Batch Size} & 8\\
\textbf{PEFT} & True\\
\textbf{Lora\_r} & 256\\
\textbf{Lora\_alpha} & 512\\
\textbf{Lora\_Dropout} & 0.1\\
\textbf{Lora Target Module} & Q proj,K proj,V proj,O proj,W proj,LM\_Head\\
\textbf{Random Seed} & 42\\
\textbf{Merging Strategy} & DARE Merging\\
\textbf{Merging p} & 0.1\\
\textbf{Merging Lambda} & 1.95\\
\textbf{Training Method} & Supervised Fine-tuning\\
\bottomrule
\end{tabular}%
\caption{\footnotesize Hyperparameters used to train \textsc{Prometheus 2} 8x7B.}
\label{tab:prometheus2_8x7b_hyperparameter}
\end{table}

\section{Quality Verification of the \textsc{Preference Collection}}\label{appendix:quality_verification}

To ensure the quality of the \textsc{Preference Collection}, particularly the generated verbal feedback ${v}_{{r}_{m},{r}_{n}}$, we employ five annotators with backgrounds in natural language processing. The annotation study was designed and administered in accordance with [Affiliation X]'s ethical guidelines. Crowd workers were informed of the potential risks of participation and researcher contact information before hand in the study consent form. The hourly wage and expected study time were informed in the Prolific platform. We compensated workers 9 GBP per hour. 3 were from USA and 2 were from Asian demographics. 

We randomly sample 200 instances with different instructions and conduct a three-part verification process. First, we assess the \textbf{coherence} of ${v}_{{r}_{m},{r}_{n}}$ with the scoring decision (\textit{i.e.}, 'A is better' or 'B is better'). Second, we evaluate the \textbf{suitability} of ${v}_{{r}_{m},{r}_{n}}$ against the evaluation criteria $e$. Lastly, to determine the \textbf{criticality} of the feedback, we compare the newly generated ${v}_{{r}_{m},{r}_{n}}$ with a concatenation of ${v}_{{r}_{m}}$ and ${v}_{{r}_{n}}$. This aims to determine if ${v}_{{r}_{m},{r}_{n}}$ effectively leverages the mutual information between ${r}_{m}$ and ${r}_{n}$. Annotators then vote on whether ${v}_{{r}_{m},{r}_{n}}$ or the concatenation of ${r}_{m}$ and ${r}_{n}$ is more critical. The results are shown in Table~\ref{tab:quality}. Note that the Preference Collection only includes English instances.

\section{Training and Inference Details}\label{appendix:hyperparameter}

The configurations we used for prompting and training evaluator LMs are shown in Table~\ref{tab:inference_hyperparameter}, \ref{tab:prometheus2_7b_hyperparameter}, \ref{tab:prometheus2_8x7b_hyperparameter}. For Auto-J, PairRM and UltraRM, we utilize their prompt template, inference hyperparameter mentioned in the model cards or github repositories in order to ensure the configuration is optimal for a fair performance comparison. For proprietary LMs, \textsc{Prometheus 1}, and \textsc{Prometheus 2} models, we use the same prompt template and evaluation configurations.

For both training and inference, we utilized eight 40GB NVIDIA A100 GPUs. Training required approximately 800 GPU hours, using the implementation from the Alignment Handbook repository\footnote{\href{https://github.com/huggingface/alignment-handbook}{https://github.com/huggingface/alignment-handbook}}. For inference, we used the vllm framework\footnote{\href{https://github.com/vllm-project/vllm}{https://github.com/vllm-project/vllm}}.

The results from Direct Assessment are averaged after three multiple runs, and pairwise grading is conducted in a single run. Instead of using error bars, we report the consistency in assessment formats, Krippendorff's alpha for consistency in direct assessment, and transitivity statistics for consistency in pairwise ranking.

\section{Direct Assessment Results: Extended}\label{appendix:direct-assessment}

Table~\ref{table:direct-assessment-kendall-tau} and \ref{table:direct-assessment-spearman} (on the next page) shows the extended results Table~\ref{table:direct-assessment}. Even when changing the metrics to either Kendall-Tau and Spearman, the overall trends are maintained. \textsc{Prometheus 2} shows superior evaluation performances among the open evaluator LMs, achieving high correlations with humans and proprietary LMs.

\section{License}

Our models are released under the Apache 2.0 license. The Preference Collection dataset is subject to OpenAI's Terms of Use for generated data. The model could be used for commercial purposes while the dataset is intended for research purposes. We used perspective API to ensure that the training data or evaluation datasets do not include PII-included instances.

\clearpage

\begin{table*}
\fontsize{9}{14}\selectfont
\centering
\resizebox{\textwidth}{!}{\begin{tabular}{lcccccccc}
    \toprule
    \multicolumn{1}{c}{\multirow{2}{*}{\textbf{Evaluator LM}}}& \multicolumn{2}{c}{\textsc{Vicuna Bench}} & \multicolumn{2}{c}{\textsc{MT Bench}} & \multicolumn{3}{c}{\textsc{FLASK}} & Feedback Bench\\ 
    \cmidrule(lr){2-3} \cmidrule(lr){4-5} \cmidrule(lr){6-8} \cmidrule(lr){9-9} & GPT-4-1106& Claude-3-Opus & GPT-4-1106& Claude-3-Opus & GPT-4-1106& Claude-3-Opus & Humans & GPT-4-0613\\
    \midrule
    \textsc{Llama2-Chat 7B}&0.183&0.203&0.065&0.070&0.229&0.186&0.211&0.419\\
    \textsc{Llama2-Chat 13B}&0.145&0.146&-0.019&0.037&0.160&0.174&0.174&0.453\\
    \textsc{Llama2-Chat 70B}&0.282&0.382&0.150&0.196&0.310&0.310&0.231&0.487\\
    \textsc{Mistral-Instruct-7B}&0.314&0.391&0.208&0.281&0.395&0.384&0.287&0.454\\
    \textsc{Mixtral-Instruct-8x7B}&0.395&0.468&\underline{0.433}&\underline{0.419}&0.410&0.408&0.304&0.551\\
    \textsc{Prometheus-7B}&0.405&0.425&0.290&0.263&0.282&0.251&0.236&0.770\\
    \textsc{Prometheus-13B}&0.397&0.434&0.299&0.352&0.365&0.352&0.299&\underline{0.793}\\
    \textsc{Auto-J (13B)}&0.282&0.242&0.303&0.272&0.312&0.282&0.312&0.515\\
    \textsc{Prometheus-2-7B}&\underline{0.543}&\underline{0.476}&0.390&0.372&\underline{0.476}& \underline{0.446}& \underline{0.377}&0.784\\
    \textsc{Prometheus-2-8x7B}&\textbf{0.559}&\textbf{0.515}&\textbf{0.535}&\textbf{0.483}&\textbf{0.526}& \textbf{0.507} & \textbf{0.388}&\textbf{0.800}\\
    \midrule
    \textsc{GPT-3.5-Turbo-0613}&0.255&0.287&0.148&0.157&0.360&0.315&0.298&0.489\\
    \textsc{GPT-4-1106}&/&0.553&/&0.590&/& 0.609 & 0.517&0.662\\
    \textsc{Claude-3-Opus}&0.553&/&0.590&/&0.609&/&0.453&0.693\\
    \bottomrule    
\end{tabular}}
\caption{\footnotesize Kendall-Tau correlations between reference evaluators (listed on top) and evaluator LMs. The best comparable statistics are \textbf{bolded} and second best \underline{underlined} except proprietary LMs.}
\label{table:direct-assessment-kendall-tau}
\end{table*}

\begin{table*}
\fontsize{9}{14}\selectfont
\centering
\resizebox{\textwidth}{!}{\begin{tabular}{lcccccccc}
    \toprule
    \multicolumn{1}{c}{\multirow{2}{*}{\textbf{Evaluator LM}}}& \multicolumn{2}{c}{\textsc{Vicuna Bench}} & \multicolumn{2}{c}{\textsc{MT Bench}} & \multicolumn{3}{c}{\textsc{FLASK}} & Feedback Bench\\ 
    \cmidrule(lr){2-3} \cmidrule(lr){4-5} \cmidrule(lr){6-8} \cmidrule(lr){9-9} & GPT-4-1106& Claude-3-Opus & GPT-4-1106& Claude-3-Opus & GPT-4-1106& Claude-3-Opus & Humans & GPT-4-0613\\
    \midrule
    \textsc{Llama2-Chat 7B}&0.236&0.255&0.084&0.089&0.301&0.244&0.279&0.511\\
    \textsc{Llama2-Chat 13B}&0.178&0.179&-0.025&0.044&0.206&0.222&0.224&0.543\\
    \textsc{Llama2-Chat 70B}&0.348&0.466&0.197&0.252&0.391&0.389&0.298&0.585\\
    \textsc{Mistral-Instruct-7B}&0.389&0.480&0.266&0.358&0.499&0.478&0.374&0.563\\
    \textsc{Mixtral-Instruct-8x7B}&0.476&0.556&\underline{0.545}&\underline{0.517}&0.505&0.500&0.386&0.659\\
    \textsc{Prometheus-7B}&0.508&0.528&0.385&0.349&0.367&0.326&0.317&0.876\\
    \textsc{Prometheus-13B}&0.492&0.534&0.401&0.470&0.474&0.454&0.398&\underline{0.893}\\
    \textsc{Auto-J (13B)}&0.337&0.297&0.408&0.365&0.402&0.358&0.408&0.623\\
    \textsc{Prometheus-2-7B}&\textbf{0.664}&\underline{0.591}&0.509&0.482&\underline{0.597}& \underline{0.555}& \underline{0.491}&0.885\\
    \textsc{Prometheus-2-8x7B}&\underline{0.660}&\textbf{0.615}&\textbf{0.669}&\textbf{0.605}&\textbf{0.642}& \textbf{0.618} & \textbf{0.496}&\textbf{0.912}\\
    \midrule
    \textsc{GPT-3.5-Turbo-0613}&0.319&0.354&0.192&0.198&0.446&0.390&0.374&0.565\\
    \textsc{GPT-4-1106}&/&0.659&/&0.721&/& 0.729 & 0.650&0.753\\
    \textsc{Claude-3-Opus}&0.659&/&0.721&/&0.729&/&0.567&0.784\\
    \bottomrule    
\end{tabular}}
\caption{\footnotesize Spearman correlations between reference evaluators (listed on top) and evaluator LMs. The best comparable statistics are \textbf{bolded} and second best \underline{underlined} except proprietary LMs.}
\label{table:direct-assessment-spearman}
\end{table*}

\begin{table*}
\centering
\fontsize{6}{8}\selectfont
\resizebox{\textwidth}{!}{\begin{tabular}{lccccccccccccc}
    \toprule
    \multicolumn{1}{c}{\multirow{2}{*}{\textbf{Evaluator LM}}}& \multicolumn{3}{c}{\textsc{HHH Alignment}} & \multicolumn{3}{c}{\textsc{MT Bench Human Judg.}} & \multicolumn{3}{c}{\textsc{Auto-J Eval}}\\ 
    \cmidrule(lr){2-4} \cmidrule(lr){5-7} \cmidrule(lr){8-10} & Direct2Pair($\uparrow$) & Pair2Pair($\uparrow$) & $\Delta$($\downarrow$) & Direct2Pair($\uparrow$) & Pair2Pair($\uparrow$) & $\Delta$($\downarrow$) & Direct2Pair($\uparrow$) & Pair2Pair($\uparrow$) & $\Delta$($\downarrow$)\\ 
    \midrule
    \textsc{Auto-J (13B)}&46.61&\underline{75.57}&28.96&48.14& 69.12 & 20.98&47.40& \underline{76.64}&29.24\\
    \textsc{Prometheus-2-7B}&\underline{74.21}&74.66&\textbf{0.45}&\textbf{63.24}&\underline{70.78}&\textbf{7.54}&\textbf{68.11}&75.07&\textbf{6.96}\\
    \textsc{Prometheus-2-8x7B}&\textbf{81.45}&\textbf{85.52}&\underline{4.07}&\underline{61.67}&\textbf{71.96}&\underline{10.29}&\underline{66.54}&\textbf{79.98}&\underline{13.44}\\
    \midrule
    \textsc{GPT-4-1106-Preview}&83.71&90.95&7.24&68.04&79.90&11.86&54.27&83.12&28.85\\
    \textsc{Claude-3-Opus}&84.62&94.57&9.95&62.65&77.65&15.00&61.04&82.90&21.86\\
    \bottomrule    
\end{tabular}}
\caption{\footnotesize \textbf{Consistency across Evaluation Formats} Pairwise ranking accuracy when assessing in direct assessment formats (denoted as `Direct2Pair') and pairwise ranking formats (denoted as `Pair2Pair'). Smaller $\Delta$ values indicate that evaluator LMs can robustly evaluate across the two different formats.}
\label{table:relative_diff}
\end{table*}  

\clearpage

\begin{table}[H]
\centering
\fontsize{7}{9}\selectfont
{
\begin{tabular}{lccc}
\toprule
\textbf{Evaluator LM} & Vicuna Ben. & MT Ben. & FLASK\\
    \midrule
    \textsc{Llama2-Chat 7B}  & 0.3558 & 0.2565 & 0.4379\\
    \textsc{Llama2-Chat 13B} & 0.2017 & 0.2998 & 0.4038\\
    \textsc{Llama2-Chat 70B} & 0.5212 & 0.4559 & 0.6204\\
    \textsc{Mistral-Instruct-7B} & 0.5157 & 0.4393 & 0.5884 \\
    \textsc{Mixtral-Instruct-8x7B} & 0.5459 & \underline{0.6229} & \underline{0.6976} \\
    \textsc{Prometheus-7B} & \underline{0.6049} & 0.5363 & 0.5970 \\
    \textsc{Prometheus-13B} & 0.5734 & 0.5181 & 0.5624 \\
    \textsc{Auto-J (13B)} & 0.4976 & 0.5069 & 0.6160 \\
    \textsc{Prometheus-2-7B} & 0.6018 & 0.5340 & 0.5991 \\
    \textsc{Prometheus-2-8x7B} & \textbf{0.6383} & \textbf{0.6862} & \textbf{0.7874} \\
    \midrule
    \textsc{GPT-3.5-Turbo-0613} & 0.7108 & 0.4800 & 0.6389 \\
    \textsc{GPT-4-1106-preview} & 0.7366 & 0.8271 & 0.8355 \\
    \textsc{Claude-3-Opus} & 0.8284 & 0.8601 & 0.8976\\
    \bottomrule    
\end{tabular}}
\caption{\footnotesize Krippendorff's alpha statistics for evaluator LMs when prompted 3 times via non-deterministic decoding.}
\label{table:krippendorff}
\end{table}

\begin{table}[H]
\centering
\fontsize{7}{9}\selectfont
\begin{tabular}{lc}
    \toprule
    \multicolumn{1}{c}{\multirow{2}{*}{\textbf{Evaluator LM}}}& \textsc{Preference Collection}\\ 
    \cmidrule(lr){2-2} & Transitivity\\ 
    \midrule
    \textsc{Mistral-Instruct-7B}&87.10\\
    \textsc{Mixtral-Instruct-8x7B}&90.45\\
    \textsc{Pair RM}&91.40\\
    \textsc{Ultra RM}&94.25\\
    \textsc{Auto-J (13B)}&89.65\\
    \textsc{Prometheus-2-7B}&\textbf{97.60}\\
    \textsc{Prometheus-2-8x7B}&\underline{96.75}\\
    \midrule
    \textsc{GPT-3.5-Turbo-0613}&84.35\\
    \textsc{GPT-4-1106-preview}&95.70\\
    \textsc{Claude-3-Opus} &96.20\\
    \bottomrule    
\end{tabular}
\caption{\footnotesize Transitivity statistics to measure consistency in pairwise ranking evaluation settings.}
\label{table:transitivity}
\end{table}

\section{Consistency of Evaluator LMs}\label{appendix:consistency}

In addition to obtaining high correlation and accuracy, achieving high consistency is another important aspect for evaluator LMs. We first test if evaluator LMs could give consistent scoring decisions in direct assessment formats. We inferencing multiple times with non-deterministic decoding (\textit{e.g.}, using temperature 1.0). Following the experimental design from \citet{ye2023flask}, we choose to inference 3 times and report the Krippendorff's alpha value. As shown in Table~\ref{table:krippendorff}, the results indicate that training on feedback data only slightly improves consistency. On the other hand, we find that the LMs with a large number of parameters achieve high consistency. This indicates the importance of selecting a large LM as the base model when training an evaluator LM. Notably, \textsc{Prometheus-2-8x7B} obtains the highest correlation among open evaluator LMs. 

Moreover, to evaluate consistency in pairwise ranking settings (Table~\ref{table:transitivity}), we measure transitivity (\textit{i.e.}, a higher score for response B over A, and for C over B, results in a higher score for C over A). As shown in Table~\ref{table:transitivity}, the \textsc{Prometheus 2} models achieve performances on par with GPT-4, showing that they could provide robust judgments in pairwise ranking schemes.

Lastly, we conduct an experiment to test if evaluator LMs could achieve consistent scores across different evaluation formats. To do this, we use pairwise ranking benchmarks and measure the performance differences when prompted with direct assessment formats and pairwise ranking formats. Specifically, following \citet{kim2023prometheus}, to process pairwise ranking datasets in a direct assessment scheme, we evaluate each response separately and compare the scoring decisions. We mark it as correct if the evaluator LM provides a higher score for the human-chosen response over the rejected one. As shown in Table~\ref{table:relative_diff} (on the previous page), the results show that \textsc{Prometheus 2} models show lower performance differences across evaluation formats, indicating their robustness.

\clearpage

\begin{table*}[t!]
\fontsize{7}{9}\selectfont
\centering
\resizebox{0.9\textwidth}{!}{\begin{tabular}{lcccccc}
\toprule
\multicolumn{1}{c}{\multirow{2}{*}{\textbf{Evaluator LM}}} & \multicolumn{3}{c}{\textsc{BiGGen Bench}}  & \multicolumn{3}{c}{\textsc{FLASK}} \\ 
\cmidrule(lr){2-4} \cmidrule(lr){5-7} & Reference-free  & Reference-based  & $\Delta$  & Reference-free  & Reference-based  & $\Delta$ \\
\midrule
\textsc{Mistral-Instruct} &0.305 &0.310 & 0.005 & 0.331 & 0.374 & 0.043 \\
\textsc{Mixtral-Instruct} &0.320 &0.322 & 0.002 & 0.377 & 0.386 & 0.009\\
\textsc{Prometheus-2-7B} &0.403 &0.455 & 0.052 & 0.425 & 0.545 & 0.120 \\
\textsc{Prometheus-2-8x7B} &0.424 &0.472 & 0.048 & 0.411 & 0.555 & 0.144 \\
\midrule
\textsc{GPT-3.5-Turbo-0613} & 0.236 & 0.252 & 0.016 &0.354 & 0.374 & 0.020 \\
\textsc{GPT-4-1106} & 0.554 & 0.599 & 0.045 &0.616 & 0.679 & 0.063 \\
\bottomrule
\end{tabular}}
\caption{\footnotesize Pearson correlations between different evaluator models with and without the reference answer and Human. Reference-based evaluations outperform reference-free evaluations across all evaluator LMs.}
\label{tab:ref_no_ref}
\end{table*}


\begin{table*}[t!]
\centering
\fontsize{8}{12}\selectfont
\resizebox{\textwidth}{!}{
\begin{tabular}{lccccccccccc}
    \toprule
    \multicolumn{1}{c}{\multirow{2}{*}{\textbf{Merging Method}}} & \multicolumn{5}{c}{\textsc{Direct Assessment Benchmarks}} & \multicolumn{5}{c}{\textsc{Pairwise Ranking Benchmarks}} & \multirow{2}{*}{Average} \\
    \cmidrule(lr){2-6} \cmidrule(lr){7-11}
    & \textsc{Vicuna Ben.} & \textsc{MT Ben.} & \textsc{FLASK (Human)} & Feedback Ben. & Average & \textsc{HHH Align.} & \textsc{MT Ben. H.J.} & \textsc{Auto-J} & Pref. Ben. & Average & \\
    \midrule
    \textsc{Linear} & 0.642 & 0.543 & \underline{0.544} & 0.878 & 0.652 & 78.73 & 67.25 & 73.80 & 92.45 & 78.06 & 82.93 \\
    \textsc{Slerp} & 0.648 & 0.532 & 0.536 & 0.879 & 0.649 & 74.66 & 70.2 & 72.33 & 92.60 & 77.44 & 82.67 \\
    \textsc{Task Arithmetic} & 0.518 & 0.497 & 0.482 & 0.831 & 0.582 & \textbf{80.09} & \underline{69.80} & 72.82 & 93.00 &  \textbf{78.93} & 81.01 \\
    \textsc{TIES} & 0.534 &  \textbf{0.567} & 0.529 & 0.826 & 0.614 & \underline{79.64} & 67.75 & 72.91 &  \textbf{93.95} & \underline{78.56} & 80.58 \\
    \textsc{Dare-Ties} & \underline{0.653} & 0.545 & 0.543 & \underline{0.880} & \underline{0.655} & \underline{79.64} & 66.57 & \underline{74.68} & \underline{93.30} & 78.55 & \underline{83.27} \\
    \textsc{Dare-Linear} & \textbf{0.666} & \underline{0.548} &  \textbf{0.545} &  \textbf{0.882} &  \textbf{0.660} & 74.66 &  \textbf{70.78} &  \textbf{75.07} & 93.25 & 78.44 &  \textbf{83.32} \\
    \bottomrule
\end{tabular}}
\caption{\footnotesize Pearson correlations and accuracy measurements across various benchmarks for different merging methods. The best comparable statistics are \textbf{bolded} and second best \underline{underlined}.}
\label{tab:merging_ablation}
\end{table*}

\clearpage



\section{Reference-free Evaluation in Direct Assessment Formats}\label{appendix:reference_free}

In this section, we assess the impact of excluding a reference answer in evaluations conducted using direct assessment formats. The results are presented in Table~\ref{tab:ref_no_ref} (on the previous page). For this experiment, we employ FLASK~\citep{ye2023flask} which includes human judgments and additionally the BiGGen Bench~\citep{kim2024biggen}. The BiGGen Bench is a generation benchmark which includes a evaluation criteria tailored to each instance and provides 2840 human judgments (excluding the multilingual tasks) in direct assessment formats.

Across both benchmarks and different evaluator LM variants, the correlation with humans diminishes when the reference answer is discarded. Even for GPT-4-1106, there is a significant performance degradation (0.045, 0.063). This suggests that including a reference answer is crucial for conducting effective evaluations with LMs. Interestingly, \textsc{Prometheus-2-7B} achieves better performance in a reference-free setting (0.403, 0.425) than Mistral-7B-Instruct-v0.2 (0.310, 0.374). Similar trends are observed for \textsc{Prometheus-2-8x7B} (0.424, 0.411) and Mixtral-8x7B-Instruct-v0.1 (0.322, 0.386). This implies that one effect of training an evaluator LM with a reference answer included is to induce the ability to ground judgments to the given reference answer.

\section{Merging Method Ablation}\label{appendix:merging}

In this section, in addition to \textbf{linear merging}, we also test different merging techniques including:

\begin{itemize}
    \item \textbf{Slerp merging}~\citep{goddard2024arcee} operates by interpolating two weights ${\theta}_{d}$ and ${\theta}_{p}$ while preserving the geometric properties of the spherical space in which ${\theta}_{d}$ and ${\theta}_{p}$ reside. Specifically, this is conducted by normalizing ${\theta}_{d}$ and ${\theta}_{p}$ into unit length and then merging the two weights based on the coefficient $\alpha$ such as: 
    \begin{equation}
    {\theta}_{final} = \alpha \times \frac{{\theta}_{d}}{||{\theta}_{d}||} + (1 - \alpha) \times \frac{{\theta}_{p}}{||{\theta}_{p}||}
\end{equation}
    \item \textbf{Task Arithmetic merging}~\citep{ilharco2022editing} which can be expressed as follows:
    \begin{equation}
\begin{aligned}
    {\theta}_{final} = {\theta}_{init} + \alpha \times ({\theta}_{d} - {\theta}_{init}) +\\ (1 - \alpha) \times ({\theta}_{p} - {\theta}_{init})
\end{aligned}
\end{equation}
where ${\theta}_{init}$ is the weight of the base model. However, we empirically find that the resulting evaluator LM ${\theta}_{final}$ often does not generate valid scoring decisions (\textit{e.g.}, generating an integer during pairwise ranking).
    \item \textbf{TIES merging}~\citep{yadav2024ties}, while similar to Task Arithmetic merging, adds (1) a \texttt{Trim} operation to remove redundant weights in the task vector ${\theta}_{d} - {\theta}_{init}$ and ${\theta}_{p} - {\theta}_{init}$ and (2) \texttt{Elect} and \texttt{Disjoint} operations to resolve disagreement (\textit{i.e.}, opposite directed weights) between ${\theta}_{d} - {\theta}_{init}$ and ${\theta}_{p} - {\theta}_{init}$.
    \item \textbf{DARE merging}~\citep{yu2023language}, while also similar to Task Arithmetic and TIES merging, performs a \texttt{Random Drop} and \texttt{Re-scale} operations in the task vector ${\theta}_{d} - {\theta}_{init}$ and ${\theta}_{p} - {\theta}_{init}$ to remove redundant weights. We find that DARE merging work best when we choose Mixtral-8x7B as our base model. \textbf{DARE-linear merging} is what was originally proposed by \citet{yu2023language}. In \textbf{DARE-TIES merging}, the \texttt{Elect} operation from \citet{yadav2024ties} is additionally added after the \texttt{Re-scale} operation.
\end{itemize}

We conduct our experiments based on the implementation from MergeKit~\citep{goddard2024arcee}.~\footnote{\href{https://github.com/arcee-ai/mergekit}{https://github.com/arcee-ai/mergekit}}

In Table~\ref{tab:merging_ablation} (on the previous page), we measure the performance of evaluator LMs employing different merging methods. In direct assessment benchmarks, DARE-Linear achieves the best performance, followed by DARE-TIES and Linear merging. In pairwise ranking benchmarks, Task Arithmetics achieves the best performance, with only a minimal difference compared to other methods. On average, DARE-Linear performs best. Based on these results, we have trained Prometheus-2-7B with DARE-Linear merging. We also opted to train Prometheus-2-8x7B using DARE-Linear merging. Although the optimal merging method might differ, we have not conducted additional experiments due to computational limitations. Future work could explore whether these findings hold true.



\clearpage

\section{\textsc{Preference Collection} Augmentation Prompt}\label{appendix:augmentation_prompt}

\begin{mybox}{Prompt for Generating Verbal Feedback in Pairwise Ranking}
\#\#\#Task Description:

An instruction (might include an Input inside it), two responses to evaluate (denoted as Response A and Response B), a reference answer, and a score rubric representing a evaluation criteria are given.

1. Write a detailed feedback explaining why \{sub\_str\}, focusing strictly on the aspects highlighted in the evaluation criteria.

2. While writing the feedback, make comparisons between Response A, Response B, and Reference Answer. Instead of examining Response A and Response B separately, go straight to the point and mention about the commonalities and differences between them.

3. While writing the feedback, do not start by mentioning \{sub\_str\} in the first sentence. Instead, try to write a reasoning process that delves into the commonalities and differences of the two responses and mention \{sub\_str\} at the last part of your justification.

4. Within the feedback, do not explicitly mention about the reference answer. For instance, do not use phrases like "Compared to the reference answer". Assume that you inherently know the reference answer which could be used to determine details that are not present in both responses under assessment.

5. Please do not generate any other opening, closing, and explanations. Just write the feedback.

6. Within the feedback, generate a string phrase "[END]" after you are finished.

\#\#\#Instruction: \{instruction\}

\#\#\#Response A: \{response\_A\}

\#\#\#Response B: \{response\_B\}

\#\#\#Reference Answer: \{reference\_answer\}

\#\#\#Score Rubric: \{criteria\}

\#\#\#Feedback: 
\end{mybox}

\section{Direct Assessment Prompt}\label{appendix:direct_assessment_prompt}

\begin{mybox}{Direct Assessment System Prompt}
You are a fair judge assistant tasked with providing clear, objective feedback based on specific criteria, ensuring each assessment reflects the absolute standards set for performance.
\end{mybox}

\begin{mybox}{Direct Assessment Prompt Template}
\#\#\#Task Description:

An instruction (might include an Input inside it), a response to evaluate, and a score rubric representing a evaluation criteria are given.

1. Write a detailed feedback that assess the quality of the response strictly based on the given score rubric, not evaluating in general.

2. After writing a feedback, write a score that is an integer between 1 and 5. You should refer to the score rubric.

3. The output format should look as follows: "Feedback: (write a feedback for criteria) [RESULT] (an integer number between 1 and 5)"

4. Please do not generate any other opening, closing, and explanations.

\#\#\#The instruction to evaluate:

\{orig\_instruction\}

\#\#\#Response to evaluate:

\{orig\_response\}

\#\#\#Score Rubrics:

\{score\_rubric\}

\#\#\#Feedback: 
\end{mybox}

\section{Pairwise Ranking Prompt}\label{appendix:pairwise_ranking_prompt}

\begin{mybox}{Pairwise Ranking System Prompt}
You are a fair judge assistant assigned to deliver insightful feedback that compares individual performances, highlighting how each stands relative to others within the same cohort.
\end{mybox}

\begin{mybox}{Pairwise Ranking Prompt Template}
\#\#\#Task Description:

An instruction (might include an Input inside it), a response to evaluate, and a score rubric representing a evaluation criteria are given.

1. Write a detailed feedback that assess the quality of two responses strictly based on the given score rubric, not evaluating in general.

2. After writing a feedback, choose a better response between Response A and Response B. You should refer to the score rubric.

3. The output format should look as follows: "Feedback: (write a feedback for criteria) [RESULT] (A or B)"

4. Please do not generate any other opening, closing, and explanations.

\#\#\#Instruction:

\{orig\_instruction\}

\#\#\#Response A:

\{response\_A\}

\#\#\#Response B:

\{response\_B\}

\#\#\#Score Rubric:

\{score\_rubric\}

\#\#\#Feedback:
\end{mybox}

\end{document}